\documentclass[conference]{IEEEtran}
\IEEEoverridecommandlockouts
\usepackage{cite}
\usepackage{amsmath,amssymb,amsfonts}
\usepackage{algorithmic}
\usepackage{graphicx}
\usepackage{textcomp}
\usepackage{xcolor}
\usepackage{algorithm}
\usepackage{algorithmic}
\usepackage{color}
\newcommand{\re}[1]{{\color{black}{#1}}}

\newcommand{\red}[1]{{\color{black}{#1}}}
\DeclareMathOperator*{\argmin}{arg\,min}

\usepackage{multirow}
\usepackage[subrefformat=parens]{subcaption}

\def\BibTeX{{\rm B\kern-.05em{\sc i\kern-.025em b}\kern-.08em
    T\kern-.1667em\lower.7ex\hbox{E}\kern-.125emX}}
\begin{document}

\title{Personalized Federated Learning with Multi-branch Architecture
}


\author{\IEEEauthorblockN{Junki Mori*}
\IEEEauthorblockA{
\textit{NEC Corporation}\\
Kanagawa, Japan \\
junki.mori@nec.com}
\and
\IEEEauthorblockN{Tomoyuki Yoshiyama*}
\IEEEauthorblockA{
yoshi123.xyz@gmail.com}
\and
\IEEEauthorblockN{Ryo Furukawa}
\IEEEauthorblockA{
\textit{NEC Corporation}\\
Kanagawa, Japan \\
rfurukawa@nec.com}
\and
\IEEEauthorblockN{Isamu Teranishi}
\IEEEauthorblockA{
\textit{NEC Corporation}\\
Kanagawa, Japan \\
teranisi@nec.com}
}


\maketitle
\begin{abstract}
\textit{Federated learning} (FL) is a decentralized machine learning
technique that enables multiple clients to collaboratively train
models without requiring clients to reveal their raw data to each other. Although traditional FL trains a single global model with average performance among clients, statistical data heterogeneity across clients has resulted in the development of \textit{personalized FL} (PFL), which trains personalized models with good performance on each client's data. A key challenge with PFL is how to facilitate clients with similar data to collaborate more in a situation where each client has data from complex distribution and cannot determine one another’s distribution. In this paper, we propose a new PFL method (pFedMB) \red{using multi-branch architecture}, which achieves personalization by splitting each layer of a neural network into multiple branches and assigning client-specific weights to each branch. \red{We also design an aggregation method to improve the communication efficiency and the model performance, with which each branch is globally updated with weighted averaging by client-specific weights assigned to the branch.} pFedMB is simple but effective in facilitating each client to share knowledge with similar clients by adjusting the weights assigned to each branch. We experimentally show that pFedMB \re{performs better} than the state-of-the-art PFL methods using the CIFAR10 and CIFAR100 datasets.
\end{abstract}


\section{Introduction}
The success of machine learning in various domains has led to the demand for large amounts of data. However, a single organization (e.g. hospital) alone may not have sufficient data to construct a powerful machine learning model. It is difficult for such organizations to obtain \re{additional data} from outside sources due to privacy concerns. \textit{Federated learning} (FL) \cite{FL}, a decentralized machine learning technique, emerged as an efficient solution to this problem. FL enables multiple clients to collaboratively build a machine learning model without directly accessing their private data. 

Traditional FL methods that aim to train a single global model perform well when the data across clients are \textit{independent and identically distributed} (IID). However, in reality, there is statistical data heterogeneity between clients, that is, data are non-IID. For example, each hospital has a  different size of patient data and their distribution varies by region (e.g. age). Training a single global model on non-IID data degrades the performance for individual clients \cite{noniidFL, Tian}. \textit{Personalized FL} (PFL) \cite{PFL} addresses this problem by jointly training personalized models fitted to each client's data distribution.

Various PFL methods have been proposed and are mainly classified into two types on the basis of how the model is personalized to each client, i.e., local customization and similarity-based \cite{fedamp}. Local-customization methods train a global model and locally customize it to be a personalized model. Local fine-tuning is a typical local-customization method \cite{fine-tuning}. Most of such methods build a global model with equal contributions from all clients even though the data distribution between some clients may be very different, which leads to invalid global model updating. \re{To address this problem, similarity-based methods have been proposed. Earlier similarity-based methods are clustering-based methods \cite{three, cfl, ifca, fl+hc}, which jointly train a model within each cluster consisting of only similar clients. Such methods are only applicable when clients are clearly partitioned and not suitable for more complex data distributions. Newer similarity-based methods, such as FedAMP \cite{fedamp} and FedFomo \cite{fedfomo}, measure the similarity between clients and updates their personalized models by weighted combination of all clients' models on the basis of the similarities. It has been shown that these methods outperform local-customization methods in simple non-IID settings, e.g., where each client is randomly assigned 2 classes or clients are grouped in advance by data distribution. However, in our experiments, FedAMP and FedFomo as well as clustering-based methods did not work well in a more complex non-IID setting. This is because directly calculating the similarities is difficult due to the complexity of data distribution.}

\re{In this paper, we aim to solve the above problem with similarity-based methods by first extending clustering-based methods. With clustering-based methods, each client uses only a model corresponding to the cluster to which it belongs. However, in a more complex non-IID setting where clients cannot be clustered, it is better to use the weighted combination of all models. By applying this insight to the level of model structure, we propose a PFL method called \textit{personalized federated learning with multi-branch architecture} (pFedMB),} which splits each layer of a neural network into multiple branches and assigns client-specific weights to each branch. Each client can obtain a personalized model fitted to its own complex distribution by learning the optimized client-specific weights and using a weighted convex combination of all the branches as a single layer. In pFedMB, an aggregation method is designed, with which each branch is updated with weighted averaging by client-specific weights assigned to the branch. This aggregation method \re{enables} similar clients to \re{automatically} share knowledge more without directly calculating the similarities, \re{as with FedAMP and FedFomo}, because similar clients \re{will learn} similar client-specific weights, which leads to improve the communication efficiency and the model performance. Our experiment showed that pFedMB outperforms various state-of-the-art PFL methods \re{and is the most balanced method that can be applied to both simple and more complex settings.}  

\section{RELATED WORKS}
\subsection{Federated Learning}
FL was first introduced by McMahan et al. \cite{FL} as FedAvg, which is one of the most standard FL algorithms for updating models locally and constructing a global model by averaging them. It has been pointed out that FL faces several challenges \cite{Tian, FLsurvey}. One is statistical heterogeneity, i.e., non-IID data problem across clients, which causes accuracy drop and parameter divergence \cite{noniidFL, Tian}. There are two approaches to address non-IID data in FL setting. One is improving the robustness of a global model to non-IID data. The other is training personalized models for individual clients.

\subsection{Improved Federated Learning on Non-IID Data} 
Studies in this direction aim to train a single global model robust to non-IID data. For example, FedProx \cite{fedprox} adds the proximal term to the learning objective to reduce the potential parameter divergence. SCAFFOLD \cite{scaffold} introduces control variates to correct the local updates. MOON \cite{moon} uses the contrastive learning method at the model level to correct the local training of individual clients.

\subsection{Personalized Federated Learning}
Unlike the above direction, PFL trains multiple models personalized to individual clients. In this paper, we focus on this direction. There are two types of PFL methods based on how the model is personalized to each client, local customization and similarity-based methods.

Local-customization methods train a single global model by usual aggregation strategy from all clients, as with FedAvg. Personalization is obtained by the local customization of the global model. A typical local-customization method is local fine-tuning \cite{fine-tuning, Yu}, which locally updates a global model for a few steps to obtain personalized models. Similar techniques are used in meta-learning methods \cite{mamlFL, aruba, perfedavg, pfedme}, which update the local models with several gradient steps by using meta-learning such as MAML \cite{maml}. Model-mixing \cite{three, l2gd, apfl} gets personalized models by mixing the global model and the local models trained on local data of each client. Parameter decoupling \cite{fedper, lg-fedavg} achieves personalization through a customized model design. In parameter decoupling, the local private model parameters are decoupled from the global model parameters. For example, FedPer \cite{fedper} separates neural networks into base layers shared among all clients and private personalization layers. With distillation-based methods \cite{fedmd, feddf}, which utilize knowledge distillation, the soft scores are aggregated instead of local models to obtain personalized models with different model architectures.

With local-customization methods, client relationships are not taken into account when updating global models. Similarity-based methods take into account client relationships and facilitate related clients to learn similar personalized models, which produces efficient collaboration. MOCHA \cite{mocha} extends multi-task learning into an FL setting (each client is treated as a task) and captures relationships among clients. However, MOCHA is only
applicable to convex models. With clustering-based methods \cite{three, cfl, ifca, fl+hc}, it is assumed that inherent partitions among clients or data distributions exist, and they cluster these partitions to jointly train a model within each cluster. \re{In our experiments, we used IFCA \cite{ifca} as a clustering-based method.} FedFomo \cite{fedfomo} is another similarity-based method that efficiently calculates optimal weighted model combinations for each client on the basis of how much a client can benefit from another’s model. With FedAMP \cite{fedamp}, instead of a single global model, a personalized cloud model for each client is maintained in the server. The personalized cloud model of each client is updated by a weighted
convex combination of all the local client models on the basis of the calculated similarities between clients. Each client trains a personalized model to be close to their personalized cloud model.

Our proposed method pFedMB is a type of parameter decoupling methods in terms of the network architecture, but can be considered as a similarity-based method in terms of the manner of updating global models. 

\subsection{Multi-branch Neural Networks}

Our research is inspired by multi-branch convolutional networks such as conditionally parameterized convolutions (CondConv) \cite{condconv}, which enables us to increase the size and capacity of a network, while maintaining efficient inference. In a CondConv layer, a convolutional kernel for each example is computed as a weighted linear combination of multiple branches by example-dependent weight. \re{We extend this architecture to the FL setting, where}
the weight associated to each branch depends on data distribution (i.e., client), not examples.

\red{Other type of multi-branch neural networks has also been proposed\cite{branchynet, RDI-Nets}. BranchyNet \cite{branchynet} inserts early exit branches, that is, side branches in the middle of the neural network and achieves fast inference by allowing test samples to exit the network early via these branches when samples can already be inferred with high confidence. Hu et al. \cite{RDI-Nets} proposes RDI-Nets, multi-exit neural networks such as BranchyNet \cite{branchynet}, which are robust to adversarial examples.}

\subsection{\red{Federated Learning Based on Multi-branch Neural Networks}}
\red{There are some studies that propose FL methods based on multi-branch neural networks \cite{fedbranch, mfedavg, semi-hfl, fl-mbnn, fedtem}. For example, multi-branch neural networks are utilized in the setting of heterogeneous FL, where the clients participated in FL have different computation capacities \cite{fedbranch, mfedavg, semi-hfl, fl-mbnn}. FedBranch \cite{fedbranch}, MFedAvg \cite{mfedavg}, Semi-HFL \cite{semi-hfl}, and FL-MBNN \cite{fl-mbnn} divide a neural network into a series of small sub-branch models by inserting early exit branches and assign a proper sub-branch to each client according to their computation capacity. FedTEM \cite{fedtem} considers applications in which the data distribution of clients changes with time and trains a multi-branch network with shared feature extraction
layers followed by one of the specialized prediction branches allocated to clients from different modes, i.e., daytime mode or nighttime mode.

Our method differs from the above methods in two respects. First, the structure of multi-branch neural networks differs between them and ours. Our pFedMB splits each layer into multiple branches and computes a weighted linear combination of them like CondConv \cite{condconv}. On the other hand, the existing methods above \cite{fedbranch, mfedavg, semi-hfl, fl-mbnn} adopt a multi-exit structure like BranchyNet \cite{branchynet}, which does not allow superposition of branches. FedTEM \cite{fedtem} is rather more related to FedPer \cite{fedper} than pFedMB, since it only divides the final layer into multiple branches and also does not consider the superposition of branches. Second, unlike the above methods, our method is a PFL method and the first to use multi-branch neural networks in PFL.}

\section{PROBLEM FORMULATION}
In this section, we introduce the PFL setting. Consider $N$ clients $C_1,\dots, C_N$. Each client $C_i$ has private local data $D_i$ composed of $n_i$ data points sampled from data distribution $\mathcal{D}_i$. Denote the total number of data samples across clients by $n=\sum_{i=1}^N n_i$. We assume non-IID data across clients, that is, data distributions $\mathcal{D}_1,\dots, \mathcal{D}_N$ are different from each other. Moreover, we suppose that there is no grouping in data distributions and each client does not know each other's data distribution. 

In this paper, we use neural networks as machine learning models. Denote the $d$ dimensional model parameter of a neural network by $w\in\mathbb{R}^d$. The model parameter $w$ is also represented as $w=(W_1,\dots,W_L)$, where $L$ is the number of linear layers such as convolutional layers or fully-connected layers, and $W_l$ is the weight matrix of the $l$-th linear layer. Let $f_i\colon \mathbb{R}^d \rightarrow \mathbb{R}$ be the training loss function associated with the training dataset $D_i$, which maps the model parameter $w\in\mathbb{R}^d$ into the real value. Traditional FL aims to find a global model $w^*$ by solving the optimization problem $w^*=\argmin_{w\in\mathbb{R}^d}\sum_{i=1}^{N}f_i(w)$.
In contrast, PFL tries to find the optimal set of personalized model parameters $\{w_1^*,\dots, w_N^*\}=\argmin_{w_1,\dots,w_N\in\mathbb{R}^d}\sum_{i=1}^{N}f_i(w_i)$.

\section{PROPOSED METHOD}
In this section, we provide details of our proposed method, \textit{personalized federated learning with multi-branch architecture} (pFedMB). 

\subsection{Architecture}
\label{architecture}
\begin{figure}[t]
\centerline{\includegraphics[width=0.95\linewidth]{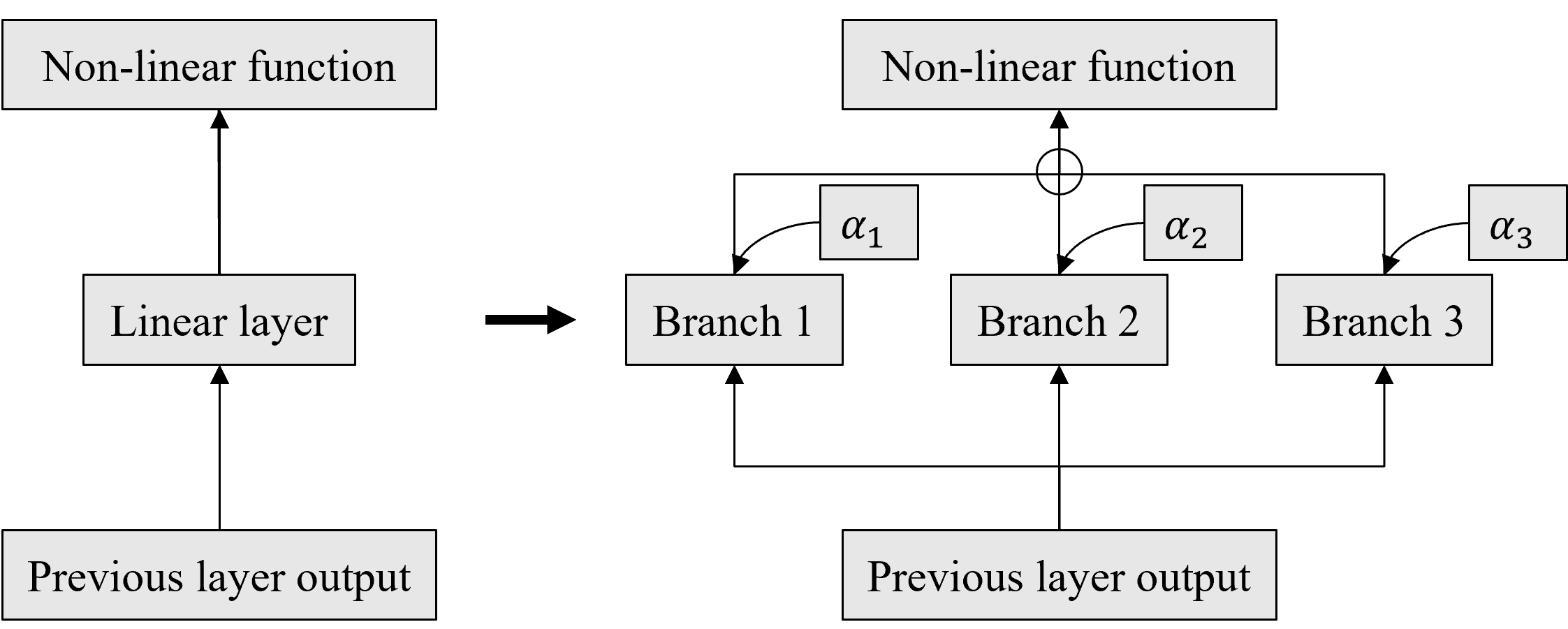}}
\caption{Architecture of multi-branch layer.}
\label{multi-branch}
\end{figure}

We first introduce the architecture of pFedMB. Our pFedMB splits each linear layer of a neural network into $B$ branches like CondConv \cite{condconv}, i.e., $W_l$ is represented as 
\begin{equation}
    \label{W_l}
    W_l=\alpha_{1,l} W_{1,l} + \dots + \alpha_{B,l} W_{B,l}, 
\end{equation}
where $W_{b,l}$ is the weight matrix of the $b$-th branch, and $\alpha_{b,l}$ is the weight assigned to the $b$-th branch satisfying $\alpha_{b,l} \geq 0$ and $\sum_{b=1}^B \alpha_{b,l} = 1$. In the multi-branch layer, each branch receives input $x$ and outputs $W_{b,l}*x$, and then the outputs of all the branches are linearly combined, weighted by $\alpha_{b,l}$. This is mathematically equivalent to the operation to multiply $x$ by $W_l$:
\begin{equation}
   W_l * x =
    \alpha_{1,l} (W_{1,l} * x) + \dots + \alpha_{B,l} (W_{B,l} * x).
\end{equation}
Therefore, $W_l$ is written as Eq.~(\ref{W_l}), which means that multi-branch layer is as  computationally efficient as normal layer in inference phase. Multi-branch architecture is shown in Fig.~\ref{multi-branch}. 

\begin{figure*}[t]
\begin{center}
\centerline{\includegraphics[width=\linewidth]{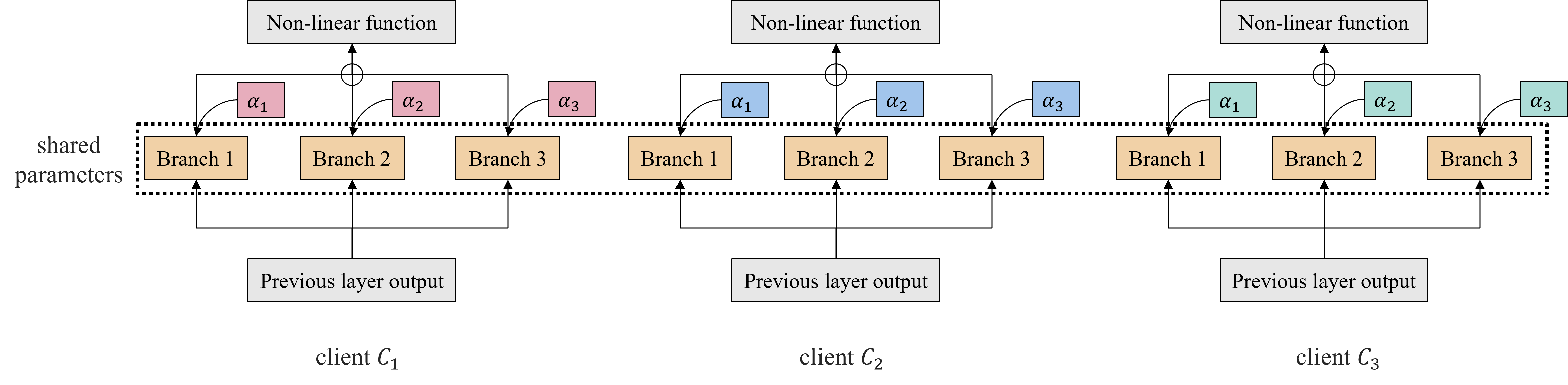}}
\caption{Architecture of pFedMB.}
\label{pFedMB}
\end{center}
\end{figure*}

In pFedMB, each client has a multi-branch neural network with the same architecture. \re{CondConv determines the weights assigned to each branch by functions of the input $x$. We extend CondConv to the PFL setting, that is, we let the weights depend on client, i.e., data distribution, not input $x$. Specifically, in order to obtain personalized models,} the weights $\{\alpha_{b,l}\}_{b,l}$ are decoupled from the global model parameters shared among all clients like FedPer \cite{fedper}, and each client locally optimizes their client-specific weights and get a model personalized to their data distribution. \re{Unlike FedPer, which decouples the parameters of the entire layer, pFedMB shares the knowledge with other clients through the parameters of each branch and decouples only the weights assigned to each branch, thus preventing overfitting on each client's data.} This architecture of pFedMB is described in Fig.~\ref{pFedMB}.

\subsection{Algorithm}

\begin{algorithm}
\caption{pFedMB: personalized federated learning with multi-branch architecture}
\label{alg:pFedMB}
\begin{algorithmic}[1]
\REQUIRE $N$ clients $\{C_i\}_{i=1}^N$, each holds $n_i$ data points, initial model parameters $W^0$ and $\{\alpha^{0,i}\}_i$,  number of communication rounds $T$, number of local epochs $E$, learning rate $\eta_{\alpha}$ and $\eta_W$, size of sampled clients $S$, number of layers $L$ and number of branches $B$
\ENSURE \ 
$N$ personalized models $w_i^T=(W^{T}, \alpha^{T,i})$
\STATE \textbf{Server do:}
\FOR {$t=0, \cdots, T-1$}
\STATE Send $W^t$ to all clients
\STATE Sample a subset of clients $\mathcal{S}_t$ with size $S$
\FOR {$C_i \in \mathcal{S}_t$ \textbf{in parallel}}
\STATE Do \textbf{ClientLocalLearning}($i$, $t$, $W^t$)
\ENDFOR
\FOR {$l=1, \cdots, L$}
\FOR {$b=1, \cdots, B$}
\STATE $W_{l,b}^{t+1} \leftarrow \frac{\sum_{i=1}^N n_i \alpha_{l,b}^{t+1,i}W_{l,b}^{t+1,i}}{\sum_{j=1}^N n_j \alpha_{l,b}^{t+1,j}}$
\ENDFOR
\ENDFOR
\ENDFOR
\STATE Return $W^T$
\STATE
\STATE \textbf{ClientLocalLearning}($i$, $t$, $W^t$):
\STATE $W^{t+1, i} \leftarrow W^t$
\STATE $\alpha^{t+1, i} \leftarrow \alpha^{t, i}$
\STATE Update $\alpha^{t+1, i}$ for $E$ epochs via Eq.~(\ref{alpha})
\STATE Update $W^{t+1, i}$ for $E$ epochs via Eq.~(\ref{W})
\STATE Return $W^{t+1,i}$ and $\alpha^{t+1,i}$ to server

\end{algorithmic}
\end{algorithm}

We review the algorithm of pFedMB in detail. Algorithm~\ref{alg:pFedMB} shows the detailed algorithm of pFedMB. We denote the model parameters of each client $C_i$ by $w_i=(W^i, \alpha^i)=(\{W_{l,b}^i\}_{l,b}, \{\alpha_{l,b}^i\}_{l,b})$ for simplicity. The algorithm is essentially based on the standard FL algorithms such as FedAvg \cite{FL}. That is, the server updates the global model at each communication round, and between rounds, each client trains their own model locally.  However, there are some differences in server and client behavior from FedAvg, respectively. Therefore, we describe those points below.

First, at each communication round $t$, the server distributes the global model parameters $W^t=\{W_{l,b}^t\}$ to each client. Local updating phase is separated into two steps, $\alpha$ updating and $W$ updating. Each client $C_i$ first updates their client-specific weights $\alpha^{t+1,i}$ based on $W^t$ for given $E$ local epochs via SGD while $W^t$ is fixed:
\begin{equation}
\label{alpha}
    \alpha^{t+1, i} \leftarrow \alpha^{t+1, i} - \eta_{\alpha} \nabla_{\alpha^{t+1,i}}f_i((W^t, \alpha^{t+1,i})),
\end{equation}
where $\eta_{\alpha}$ is the learning rate. Updating $\alpha^{t+1,i}$ allows each client $C_i$ to obtain the optimal ratio for the branches suitable for their own distribution.
Next, client $C_i$ fixes $\alpha^{t+1,i}$ and updates $W^t$ \re{for $E$ local epochs} to get the local version of the global model $W^{t+1, i}$:
\begin{equation}
\label{W}
    W^{t+1, i} \leftarrow W^{t+1, i} - \eta_{W} \nabla_{W^{t+1,i}}f_i((W^{t+1,i}, \alpha^{t+1,i})).
\end{equation}
Finally, each client sends model parameters $(\alpha^{t+1,i}, W^{t+1, i})$ to the server and the server aggregates them to obtain a new global model parameters $W^{t+1}=\{W_{l,b}^{t+1}\}_{l,b}$. In this global model updating phase, we design an $\alpha$-weighted aggregation method through the equation
\begin{equation}
\label{alpha-weight}
    W_{l,b}^{t+1} = \frac{\sum_{i=1}^N n_i \alpha_{l,b}^{t+1,i}W_{l,b}^{t+1,i}}{\sum_{j=1}^N n_j \alpha_{l,b}^{t+1,j}}.
\end{equation}
Here, $W_{l,b}^{t+1}$ is calculated through weighted averaging by not only the number of data points $n_i$, but also the client-specific weight $\alpha_{l,b}^{t+1,i}$ corresponding to $W_{l,b}^{t+1, i}$, which means that the clients who give more attention to the $b$-th branch of the $l$-th layer contribute to the calculation of $W_{l,b}^{t+1}$ more.
Therefore, this facilitates similar clients (i.e., clients who have similar $\alpha$) to collaborate more as in FedFomo \cite{fedfomo} and FedAmp \cite{fedamp}. \re{However, unlike FedFomo and FedAMP, each client only optimizes their weights assigned to each branch in pFedMB, and therefore collaboration among similar clients occurs automatically without directly calculating the similarities.}

We note that 

\section{EXPERIMENTS}
\label{experiments}
In this section, we evaluate the performance of pFedMB in non-IID data settings. First, we observe how pFedMB achieves personalization in a simple setting. Next, we compare the performance of pFedMB with the state-of-the-art PFL methods in more complex settings.

\begin{figure*}[t]
    \begin{minipage}{1\linewidth}
    \centering
    \includegraphics[width=0.7\linewidth]{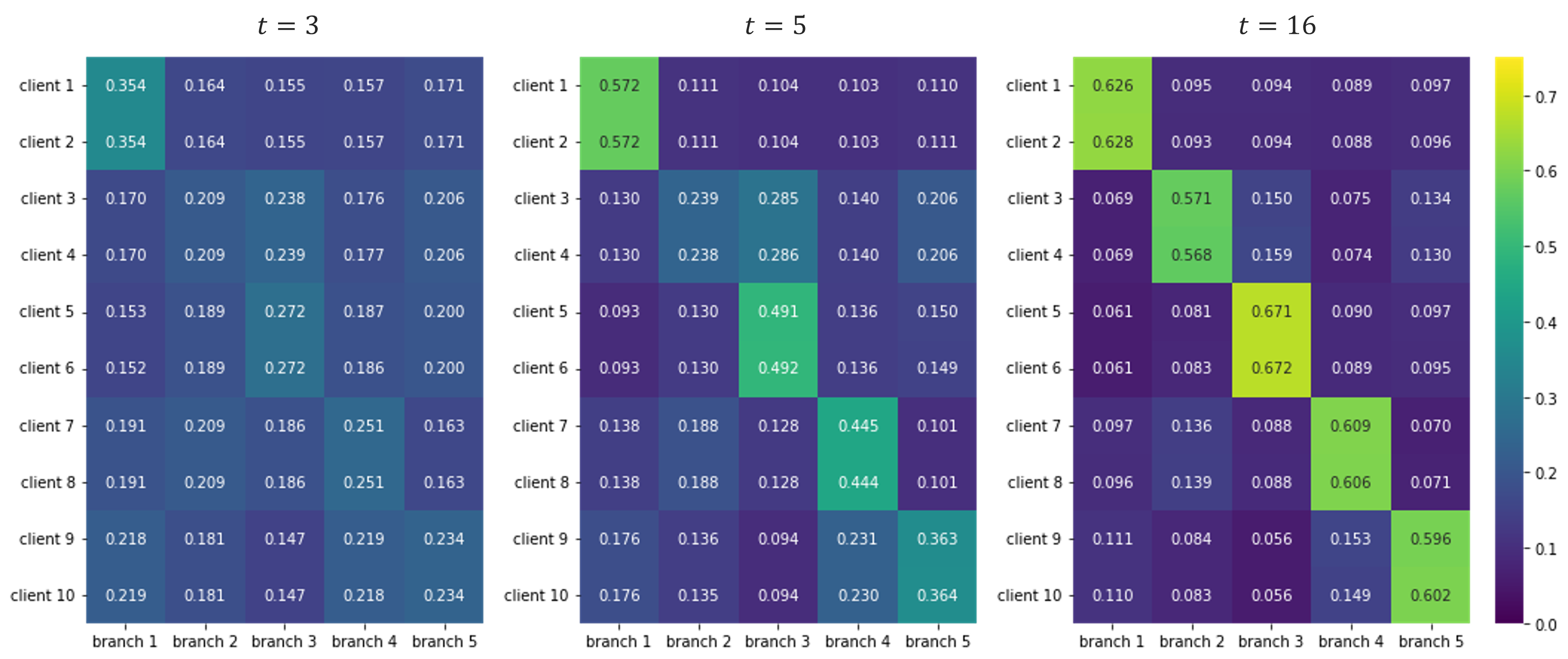}
    \subcaption{\re{pFedMB with $\alpha$-weighted averaging.}}
    \label{alpha-vis}
    \end{minipage}
    \begin{minipage}{1\linewidth}
    \centering
    \includegraphics[width=0.7\linewidth]{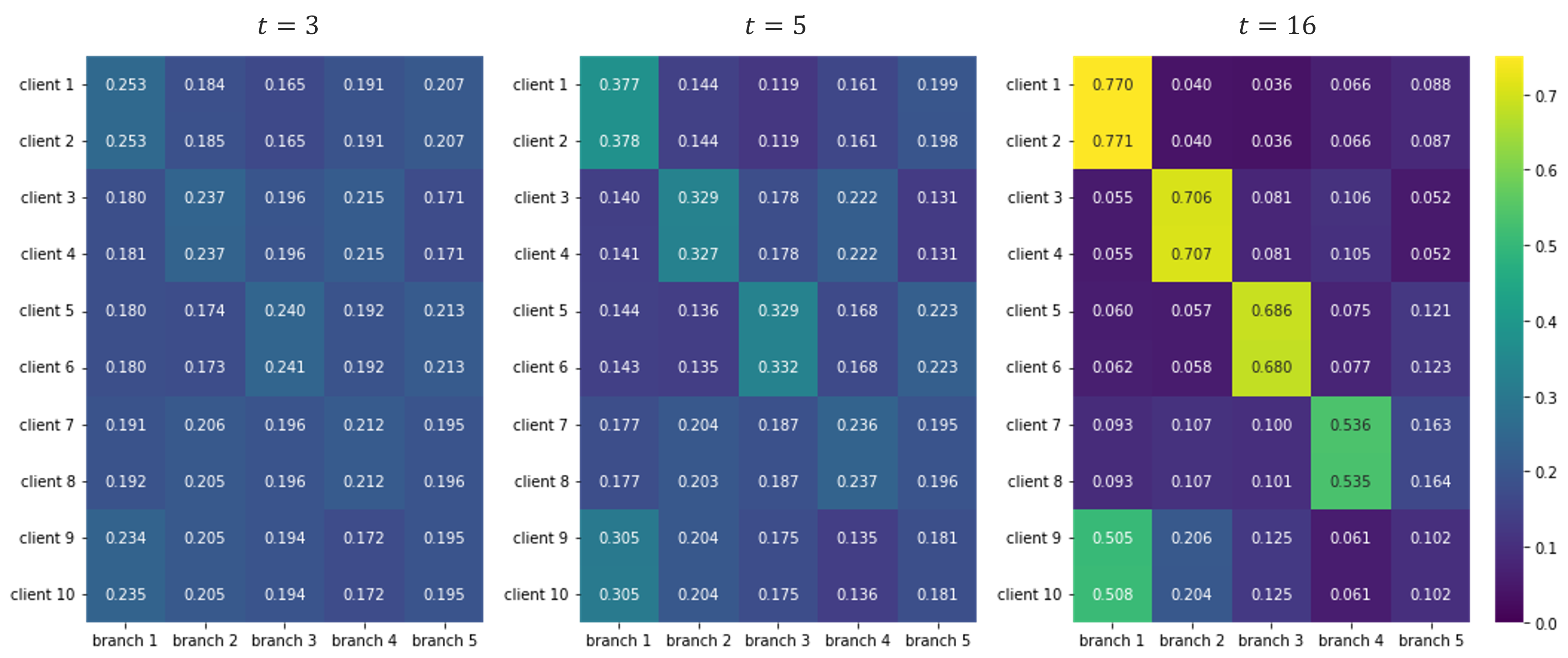}
    \subcaption{\re{pFedMB without $\alpha$-weighted averaging.}}
    \label{alpha-vis-non}
    \end{minipage}
\caption{\re{Visualization of $\alpha$ at communication round $t$ with/without $\alpha$-weighted averaging usine CIFAR10.}}
\end{figure*}


\subsection{Experimental setup}

\paragraph{Data settings} 
In this paper, we numerically evaluate the performance of pFedMB using CIFAR10 \red{and CIFAR100} \cite{cifar10}. \red{CIFAR10/CIFAR100 dataset has 50,000 training images and 10,000 test images with 10/100 classes.} We split the training images into 40,000 training images and 10,000 validation images used to tune the hyperparameters for each method (e.g. learning rate $\eta_{\alpha}, \eta_{W}$ and number of branches for pFedMB). Each type of data (i.e., training, validation, and test) is distributed across clients in the same way. For CIFAR10, we consider two different scenarios for simulating non-IID data across clients. The first is a commonly used scenario where each client is randomly assigned 2 classes \red{with the same data size}. However, this setting is not realistic and in practice each client's data is sampled from a more complex distribution. In order to simulate this, we consider another scenario where the non-IID data partition is generated by Dirichlet distribution (with concentration parameter 0.4) as in the previous studies such as \cite{moon, dirichlet, dirichlet2}.
\red{
In contrast, applying the Dirichlet distribution to CIFAR100 would result in an extremely small number of data for each class held by each client. Therefore, for CIFAR100, we adopt the non-IID data setting used in \cite{pfedme, pfedhn}, where data is heterogeneous in terms of not only classes but also data size. First, we randomly assign 20 classes for each client. Next, for each client $C_i$ and chosen class $c$ by $C_i$, we sample $u_{i, c} \in U(0.3, 0.7)$ (uniform distribution between $0.3$ and $0.7$) and allocate $\frac{u_{i,c}}{\sum_j u_{j, c}}$ of the samples for class $c$ to $C_i$.  
}

\paragraph{Implementation details} 
The performance of all the methods is evaluated by mean test accuracy among clients. Moreover, the experiments were conducted three times and the average of the test accuracies from the three experiments are reported. In each method, we train the models until the convergence. The number of clients is set to 15 and 50 in the main results and the client participation rate at each communication round is 100\% and 20\%, respectively. In all the experiments, we use a CNN network, which has two 5x5 convolution layers followed by 2x2 max pooling (the first with 6 channels and the second with 16 channels) and two fully connected layers with ReLU activation (the first with 120 units and the second with 84 units). pFedMB extends each linear layer to a multi-branch layer. The initial weights $\alpha$ for each client are set to the same value for all branches. Throughout the experiments, the number of local
epochs and batch size are set to 5 and 64, respectively.
All methods are implemented in PyTorch.

\paragraph{Compared methods}
We compare pFedMB with other various methods classified into two categories. The first category is the improved FL approach to train a single global model that performs well on non-IID data. In this category, we use FedAvg \cite{FL} and FedProx \cite{fedprox}. The second category is PFL to train multiple models personalized to each client's data distribution. We compare our method with non-federated local learning, FedPer \cite{fedper}, pFedMe \cite{pfedme}, IFCA \cite{ifca}, FedFomo \cite{fedfomo}, and FedAMP \cite{fedamp}. In all the methods, we perform local fine-tuning for local epochs. Therefore, in our experiments, FedAvg and FedProx also \re{eventually produce} multiple models.

\subsection{Process of personalization}
\label{process}
Before moving on to the main result, we will see how our method works in a simple setting. We use CIFAR10 and consider 10 clients grouped into 5 pairs, each with the same two classes for simplicity (e.g. clients 1, 2 have class 0 and 1). Furthermore, we suppose client-specific weights $\alpha$ are common for all the layers. In this experiment, the number of branches is set to 5. We compare pFedMB with local learning and FedAvg as baselines. Moreover, in order to see the effect of an aggregation method using weighted averaging by $\alpha$ (\ref{alpha-weight}) in global model updating phase, we implement pFedMB that updates global models by normal averaging as in FedAvg:
\begin{equation}
    W_{l,b}^{t+1} = \frac{\sum_{i=1}^N n_i W_{l,b}^{t+1,i}}{\sum_{j=1}^N n_j}.
\end{equation}

\begin{table}[h]
\caption{The mean test accuracy (\%) among 10 clients grouped into 5 pairs using CIFAR10. Each method includes local fine-tuning for local epochs $E$.}
\label{pre_result}
\begin{center}
\begin{tabular}{|c|c|c|c|}
\hline 
\multirow{2}{*}{Local} & \multirow{2}{*}{FedAvg} & pFedMB & \multirow{2}{*}{pFedMB} \\
 &  & (w/o $\alpha$-weighted avg.) & \\
\hline
\hline
87.26  & 87.99  & 89.08  & \textbf{89.49} \\
\hline
\end{tabular}
\end{center}
\end{table}

\begin{figure}[t]
\centerline{\includegraphics[width=0.9\linewidth]{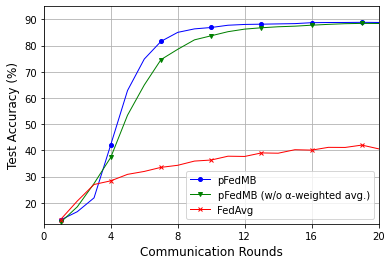}}
\caption{\red{The mean test  accuracy among 10 clients grouped into 5 pairs at each communication round using CIFAR10. Each method will be followed by local fine-tuning.}}
\label{communication-round}
\end{figure}

\begin{table*}[th]
\caption{The mean test accuracy (\%) among 15 clients and 50 clients in random 2 classes and Dirichlet distribution settings using CIFAR10. Client participation rate at each communication round is 100\% and 20\%, respectively. Each method includes local fine-tuning for local epochs $E$. ``Local'' is a non-federated local learning.}
\label{main_result}
\begin{center}
\begin{tabular}{|l|c|c|c|c|c|c|}
\hline 
\multirow{4}{*}{\textbf{Methods}} & \multicolumn{4}{|c|}{\textbf{CIFAR10}}& \multicolumn{2}{|c|}{\textbf{CIFAR100}} \\
\cline{2-7}
& 
\multicolumn{2}{|c|}{\multirow{2}{*}{\textbf{Random 2 classes}}}& \multicolumn{2}{|c|}{\multirow{2}{*}{\textbf{Dirichlet}}}  
&
\multicolumn{2}{|c|}{\textbf{Random 20 classes}} \\
& \multicolumn{2}{|c|}{} & \multicolumn{2}{|c|}{}
& \multicolumn{2}{|c|}{\textbf{with different size}}
\\
\cline{2-7}
&\  15 clients \ &\ 50 clients \ &\ 15 clients \ &\ 50 clients \ &\ 15 clients \ &\ 50 clients \ \\
\hline
\hline
Local &\ 90.01 \ &\ 83.79 \ &\ 67.54 \ &\ 61.27 \ &\ 39.10 \ &\ 29.42 \ \\
\hline
FedAvg \cite{FL} &89.69 &85.68 &73.52 &70.32 &43.49 &38.53 \\
\hline
FedProx \cite{fedprox} &90.51 &86.07 &73.38 &70.98 &44.81 &36.07 \\
\hline
FedPer \cite{fedper} &89.73  &84.56 &71.97 &67.57 &41.52 &29.67 \\
\hline
pFedMe \cite{pfedme} &90.04 &83.09 &72.92 &71.00 &\textbf{45.57} & 39.04 \\
\hline
IFCA \cite{ifca} &90.17 &85.68 &72.42 &69.84 &41.46 &35.01 \\
\hline
FedFomo \cite{fedfomo} &90.12 &85.01 &67.55 &62.62 &38.19 &28.18 \\
\hline
FedAmp \cite{fedamp} &\textbf{90.73} 	&85.90 &72.91 &65.98 &41.27 &29.53\\
\hline
pFedMB (ours) &90.69 &\textbf{86.73} &\textbf{73.53} &\textbf{72.14} &44.73 &\textbf{39.78}\\
\hline
\end{tabular}
\end{center}
\end{table*}

Table~\ref{pre_result} shows the mean test accuracy for each method. First, from Table~\ref{pre_result}, we can see that pFedMB improves FedAvg and $\alpha$-weighted averaging is effective in the perspective of accuracy. 

Next, the process of getting the optimal $\alpha$ for each client is visualized in Fig.~\ref{alpha-vis}, \ref{alpha-vis-non}, each of which is the result of pFedMB with/without $\alpha$-weighted averaging, respectively. Fig.~\ref{alpha-vis} shows that every pair of clients who have the same classes gets almost the same $\alpha$ and different pairs of clients focus on different branches. From this result, we can see that pFedMB with $\alpha$-weighted averaging achieves personalization by facilitating similar clients to collaborate more. In contrast, in pFedMB without $\alpha$-weighted averaging, the clients are clustered slowly, which results in different pairs of clients concentrating on the same branch. This leads to inefficient collaborating. \red{Fig.~\ref{communication-round} shows that $\alpha$-weighted averaging is also effective in terms of communication efficiency. Fig.~\ref{communication-round} presents the test accuracy at each communication round. pFedMB with $\alpha$-weighted averaging is clearly faster to convergence than pFedMB without $\alpha$-weighted averaging as well as FedAvg, that is, reduces communication cost.} Therefore, it is revealed that $\alpha$-weighted averaging (\ref{alpha-weight}) is effective to promote collaboration among similar clients, which leads to good performance for each client\red{ and fast convergence}.

\subsection{Performance comparison}

Table~\ref{main_result} shows the performance comparison of pFedMB and the other PFL methods in six settings. We can see that pFedMB is the best in almost all the settings. Even in 15 clients with random 2 classes setting using CIFAR10 \red{and 15 clients setting using CIFAR100}, pFedMB is comparable to the best methods FedAMP \red{and pFedMe, respectively}. We note that FedAvg and FedProx, originally not personalized FL methods also perform well especially in random 2 classes distribution setting using CIFAR10. It is because in our experiments, all the methods include local fine-tuning. That is, local fine-tuning has enough personalization power especially in a simple setting where each client has only 2 classes, which is a fact not often pointed out in the previous studies. It is also worth noting that the state-of-the-art methods such as FedAMP and FedFomo do not perform well \red{except in random 2 classes distribution setting using CIFAR10}. As mentioned in the introduction, this is probably because the similarity calculation between clients does not work well when each client's data is sampled from a complex distribution such as Dirichlet distribution \red{and random 20 classes with different size}. In contrast, pFedMB, which does not directly calculate the similarity, shows good performance in both distribution settings. In other words, pFedMB is the most balanced method that is effective in every setting.

\subsection{\red{Effects of number of branches}}
\begin{figure}[t]
\centerline{\includegraphics[width=0.9\linewidth]{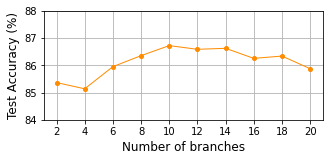}}
\caption{\red{Variation of the mean test accuracy among 50 clients in the random 2 classes distribution setting using CIFAR10 with respect to the number of branches.}}
\label{branches}
\end{figure}

\red{
We next investigate the effects of number of branches on the performance of pFedMB. Table~\ref{num_branch} shows tuned number of branches of pFedMB in each experimental setting. From Table~\ref{num_branch}, we can see that the required number of branches is at most 10. Unlike CIFAR10, CIFAR100 has a larger number of classes, and therefore the number of classes shared among clients when using CIFAR100 is larger, which results in fewer branches. Therefore, in the random 2 classes and 50 clients setting using CIFAR10, where the required branches is large, we see the effect of number of branches. In Fig.~\ref{branches}, the number of branches of pFedMB is changed from 2 to 20. The test accuracy increases with the number of branches, reaching a peak at 10 and then decreasing with the number of branches. Therefore, when using pFedMB, it is necessary to find the peak number of branches by fine-tuning.
}

\begin{table}[t]
\caption{\red{Tuned number of branches of pFedMB in each experimental setting.}}
\label{num_branch}
\begin{center}
\begin{tabular}{|c|c|c|c|}
\hline 
\multicolumn{3}{|c|}{\textbf{Settings}} & \textbf{\# of branches} \\
\hline
\hline
\multirow{4}{*}{CIFAR10} & \multirow{2}{*}{Random 2 classes} & 15 clients & 6 \\
\cline{3-4}
& & 50 clients & 10 \\
\cline{2-4}
& \multirow{2}{*}{Dirichlet} & 15 clients & 7 \\
\cline{3-4}
& & 50 clients & 6 \\
\cline{1-4}
\multirow{2}{*}{CIFAR100} & Random 20 classes & 15 clients & 3 \\
\cline{3-4}
& with different size & 50 clients & 3 \\
\hline
\end{tabular}
\end{center}
\end{table}

\section{DISCUSSIONS}

We first discuss the communication and computation costs of pFedMB. The communication and computation overheads are determined by the number of branches $B$. Let $P$ be the number of model parameters used in the traditional FL methods, e.g., FedAvg. If pFedMB splits every layer into $B$ branches, the number of model parameters increases to $B \times P$. Therefore, the communication and computation costs of pFedMB are $B$ times those of FedAvg. Those costs can be reduced by limiting the number of layers to be branched or by stopping both training and communicating the branches that are assigned small client-specific weights per clients (because these branches are not important to the corresponding client). As mentioned in Section \ref{architecture}, the computation cost of pFedMB during inference is the same as FedAvg.

Given the communication and computation bottlenecks of pFedMB, the question arises how many branches are needed. The obvious upper limit is the number of clients. This is because if the number of branches is greater than the number of clients, there is no point in collaboration among clients, and local learning is sufficient. If the clients are clustered into some groups, as in the experiment of Section~\ref{process}, then $B$ should be set to the number of groups. Also, if the data distribution of each client is a mixture of several distributions, $B$ should be set to the number of underlying distributions. If the exact number cannot be specified in advance, $B$ must be treated as a hyperparameter. In our experiments, the number of branches did not exceed 10 as a result of tuning in any setting.


We finally discuss the privacy issues for each client. Unlike FedFomo and FedAMP, pFedMB shares the clients' local models only with the central server, not with other clients or cloud server introduced in FedAMP. This is the benefit of pFedMB. However, as in clustering-based methods, there might be somewhat privacy concern for pfedMB when sending client-spcific weights $\alpha^i$ to the server because $\alpha^i$ contains the information about data distribution for client $C_i$. Note that $\alpha^i$ does not contain the information about raw data. In real-world systems, we suggest using tools such as secure multiparty computation when conducting $\alpha$-weighted aggregation if needed.

\section{CONCLUSIONS}
We proposed a PFL method called \textit{personalized federated learning with multi-branch architecture} (pFedMB), which achieves personalization by splitting each layer of a neural network into multiple branches and assigning client-specific weights to each branch. We also presented an efficient aggregation method to enable fast convergence and improve the model performance. We numerically showed that pFedMB performs better in both simple and complex data distribution settings than the other state-of-the-art PFL methods, i.e., it is the most balanced PFL method.


\bibliographystyle{IEEEtran}
\bibliography{IEEEexample}

\end{document}